# Hysteretic Behavior Simulation Based on Pyramid Neural Network: Principle, Network Architecture, Case Study and Explanation

Yongjia Xu [1], Xinzheng Lu [2*], Yifan Fei [3], Yuli Huang [2]

[1.] Zhejiang University, Zhejiang University–University of Illinois at Urbana Champaign Institute, Hangzhou, Zhejiang, China, 314400

[2.] Department of Civil Engineering, Tsinghua University, Beijing, P.R. China, 100084

[3.] Beijing Engineering Research Center of Steel and Concrete Composite Structures, Tsinghua University, Beijing, P.R. China, 100084

## Abstract

An accurate and efficient simulation of the hysteretic behavior of materials and components is essential for structural analysis. The surrogate model based on neural networks shows significant potential in balancing efficiency and accuracy. However, its serial information flow and prediction based on single-level features adversely affect the network performance. Therefore, a weighted stacked pyramid neural network architecture is proposed herein. This network establishes a pyramid architecture by introducing multi-level shortcuts to integrate features directly in the output module. In addition, a weighted stacked strategy is proposed to enhance the conventional feature fusion method. Subsequently, the redesigned architectures are compared with other commonly used

Corresponding author: Xinzheng Lu, Room No.418, New Civil Engineering Building, Department of Civil Engineering, Tsinghua University, Haidian District, Beijing, P.R. China, 100084, E-mail: luxz@tsinghua.edu.cn

network architectures. Results show that the redesigned architectures outperform the alternatives in 87.5% of cases. Meanwhile, the long and short-term memory abilities of different basic network architectures are analyzed through a specially designed experiment, which could provide valuable suggestions for network selection.

## 1. Introduction

An accurate and efficient simulation of hysteretic behavior is essential for structural (or component/material) analysis. Significant progress has been made in refined physical models (e.g., Tao and Shahsavari, 2017; Wang et al., 2020a). Meanwhile, the complexity of refined models has increased considerably, resulting in an apparent increase in the computational workload.

Many studies have focused on data-driven surrogate models to balance accuracy and efficiency. Conventional surrogate models are generally developed based on testing data and simulation results (e.g., Long and Lee, 2012; Pham et al., 2016; Liu and Guo, 2021; Lu and Guan, 2021), which exhibits certain disadvantages in terms of universality and accuracy: (1) the model simplification process relies significantly on artificially determined function forms and key assumptions or parameters; (2) for hysteretic behaviors with complex features and limited data, a simplified and accurate surrogate model is difficult to be obtained.

Deep learning has been developing rapidly in recent years. The deep neural network model involves numerous parameters, exhibits strong nonlinear fitting ability and "end-to-end" characteristics (from the input to output directly, and delivers a complete solution without artificially managing the intermediate process), effectively overcoming the disadvantages of conventional surrogate models. Therefore, deep learning methods have garnered increasing attention.

Like natural language processing, the essence of hysteretic behavior simulation is "sequence modeling." Therefore, the mainstream neural networks adopted in similar studies are recurrent neural network (RNN) (particularly the long short-term memory (LSTM) neural network (Hochreiter and Schmidhuber, 1997), gated recurrent unit (GRU) network (Cho et al., 2014)), and Transformer (or modified Transformer) (e.g., Vaswani et al., 2017; Wang and Sun, 2018; Zhang et al., 2019; Gorji et al., 2020; Xu et al., 2021). Other technologies, such as multi-layer perception (MLP) and convolutional neural networks (CNN), are also reported (e.g., Zhang et al., 2020; Lu et al., 2021).

In these networks, the forward propagation is primarily serial (i.e., from the encoder to the decoder and from the first layer to the final layer), and the prediction is conducted based on single-level features. Relevant studies in the field of computer vision (e.g., Ronneberger et al., 2015; He et al., 2016; Tan et al., 2020) show that multi-level feature fusion improves the prediction performance, and "Shortcut" is a reliable multi-level feature extraction and fusion technology. A neural network that integrates multi-level features is a “pyramid network.”

Some existing studies have added shortcuts to LSTM or Transformer networks for time-series modeling. For example, the classic Transformer architecture (Vaswani et al., 2017) involves intra-module shortcuts. Kim et al. (2017) and Zhao et al. (2018) introduce shortcuts in LSTM to improve network performance. Emelin et al. (2019) propose a lexical shortcut mechanism. Hao et al. (2019) propose a new module known as the recurrence encoder and transfer its outputs to the attention layer in the decoder through shortcuts. Liu et al. (2020) integrate the outputs of LSTM and the attention module for prediction through shortcuts.

Although existing studies have proposed and investigated the performance of many kinds of neural networks, supplementary studies are still necessary for the practical application of data-driven surrogate modeling. Shortcuts that forego multiple layers/modules have yet to receive enough attention. The feature fusion mechanism considers different levels of features equally, disregarding the variety among different levels of features. In addition, comprehensive comparison among commonly used networks and explanations of their variance in learning abilities are also valuable, while existing studies still need to catch up in these aspects.

Therefore, this study aims to enhance existing networks to improve the reliability of data-driven hysteretic behavior simulations. A weighted stacked pyramid network architecture is proposed herein, including the multi-level shortcuts and weighted stacked strategy (Section 2). Multi-scale datasets are established (Section 3). Case studies and network comparisons are carried out (Sections 4 and 5). Meanwhile, an innovative experiment is designed to compare the long and short-term memory ability of different basic network architectures (Section 6).

## 2. Methodology

### 2.1 Supporting Technology

LSTM (Hochreiter and Schmidhuber, 1997) and GRU (Cho et al., 2014) are two typical improved networks of classic RNNs widely used in sequence modeling. Meanwhile, inspired by the human attention mechanism, Bahdanau et al. (2014) propose the attention mechanism for the neural network to allocate network resources to crucial information reasonably. Vaswani et al. (2017) modify the attention mechanism and propose the Transformer network, which shows high learning ability in sequence modeling.

### 2.2 Multi-level shortcut

As mentioned above, the prediction of the LSTM and Transformer depends only on the features extracted by the final layer/module. After extraction and condensation, the feature of the final layer is the most closely associated with the output. However, because the network's learning ability is not infinite, such features could not retain all essential information (Fig. 1(a)). Therefore, an architecture with multi-level shortcuts is proposed herein. Based on conventional serial information propagation, shortcuts that directly connect the intermediate layer and output modules are added (Fig. 1(b)).

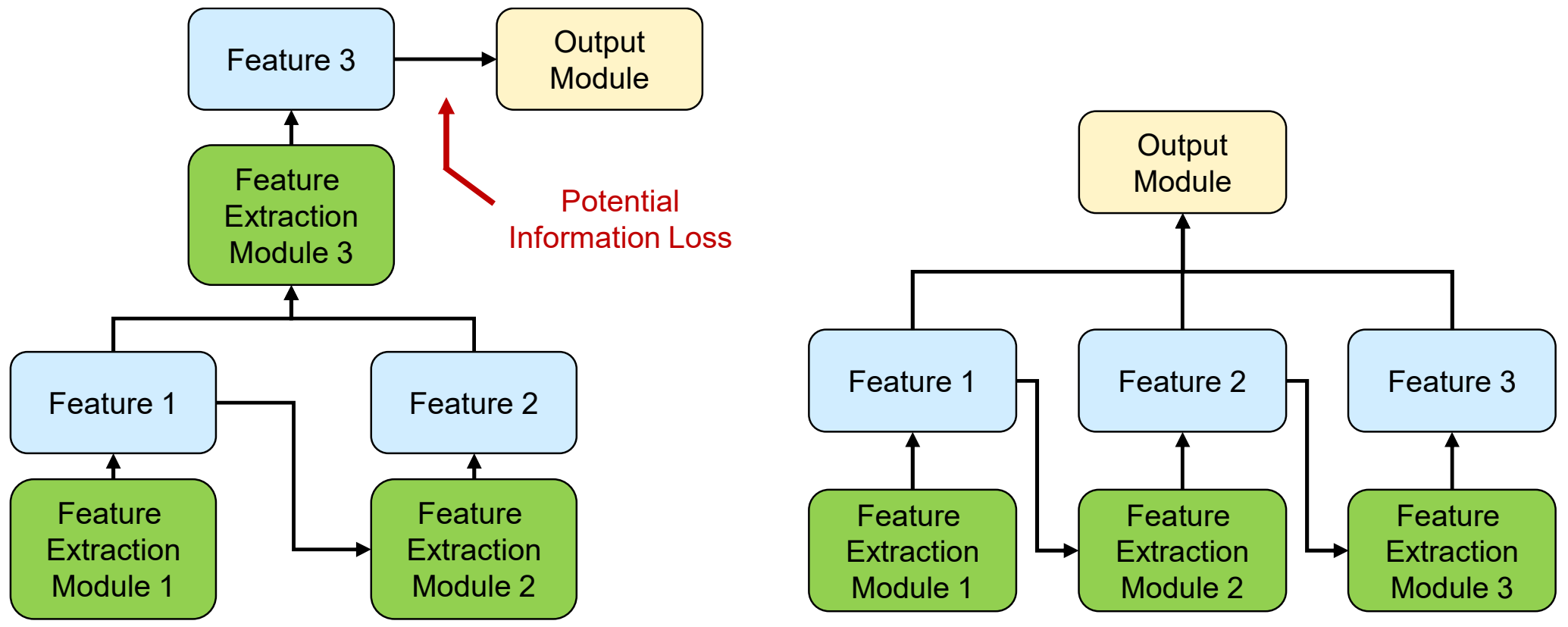

(a) Traditional shortcuts (b) The proposed multi-level shortcuts

**Fig. 1** The comparison between two addition manners of shortcuts

Integrating multi-level features allows the output module to realize final predictions based on more valuable information, thereby improving the network performance. The overall architecture of the redesigned pyramid networks with multi-level shortcuts is shown in Fig. 2(a). The concept of adding multi-level shortcuts is versatile, and the overall architecture could be maintained under different situations. Only the basic units in the networks need to be modified.

If the new architecture is redesigned based on a traditional Transformer (named "Pyramid-Transformer"), the basic unit in the network is shown in Fig. 2(b). Noting that the outputs of the encoder modules are connected to all the decoder modules. Under certain situations, the simpler network LSTM could also be used as the fundamental network, and the redesigned architecture is named “Pyramid-LSTM.” Pyramid-LSTM uses the same unit as LSTM cells (Hochreiter and Schmidhuber, 1997). Thus, the encoder and decoder sections will be identical. Furthermore, some existing studies have combined the LSTM/GRU and attention mechanism for hysteretic behavior simulation (e.g., Wang et al., 2020b; Li et al., 2021a, 2021b). The basic unit in the GRU + attention (GA) network is similar to that in Pyramid-Transformer, while a GRU layer is added in each unit, as shown in Fig. 2(c). The redesigned network is named “Pyramid-GA.”

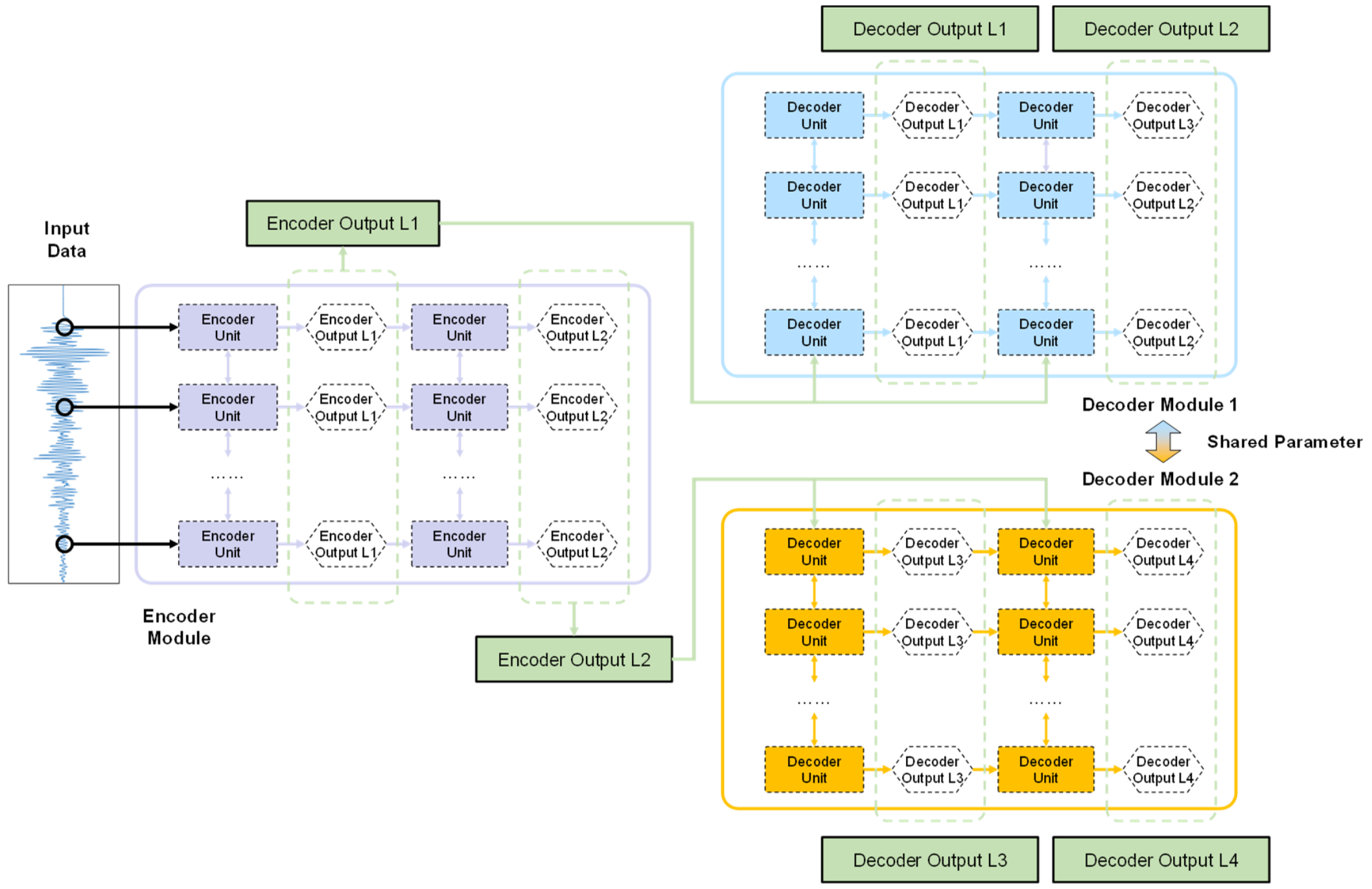

(a) The proposed multi-level shortcut

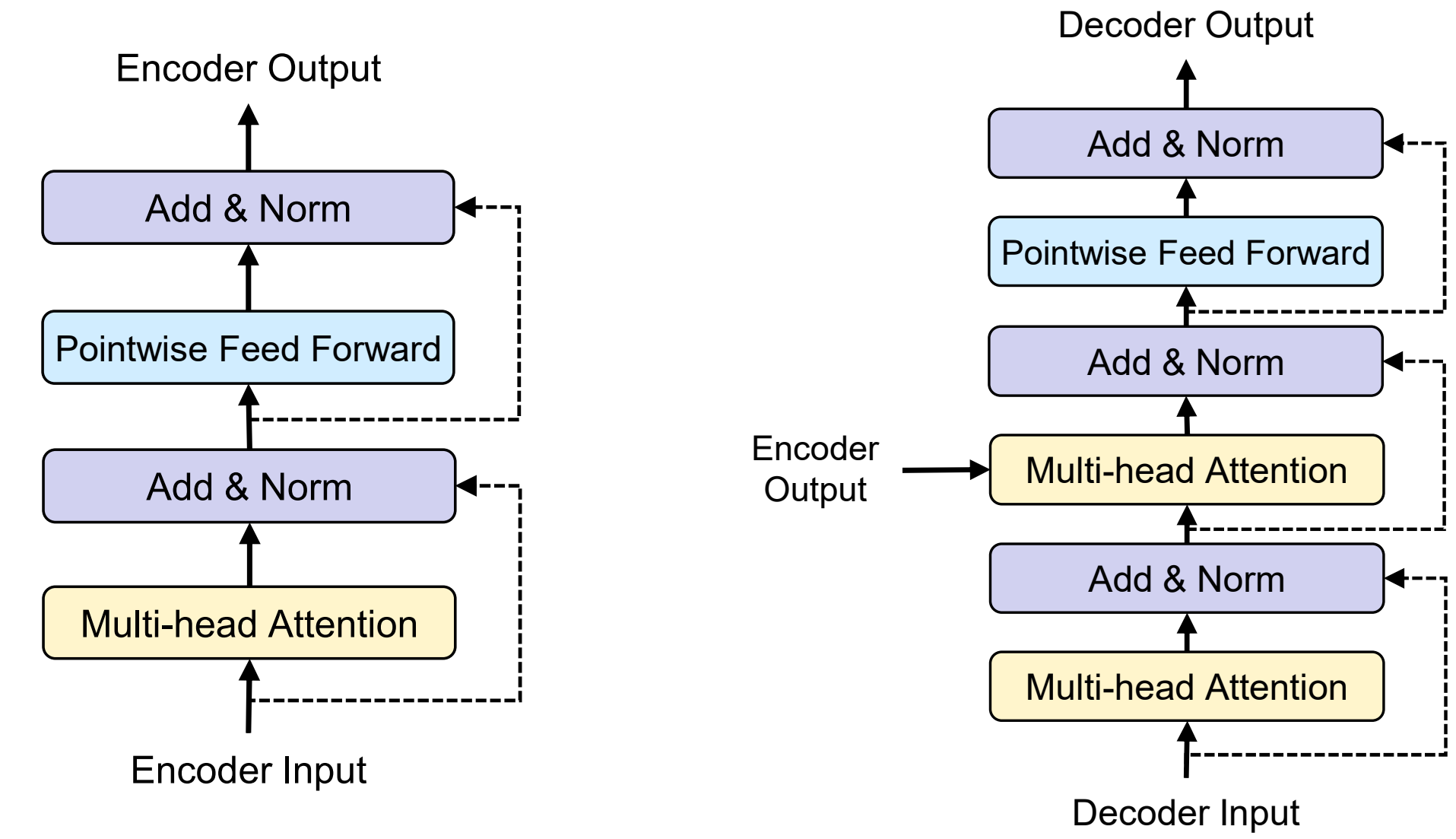

(b) Encoder and Decoder Unit in Transformer

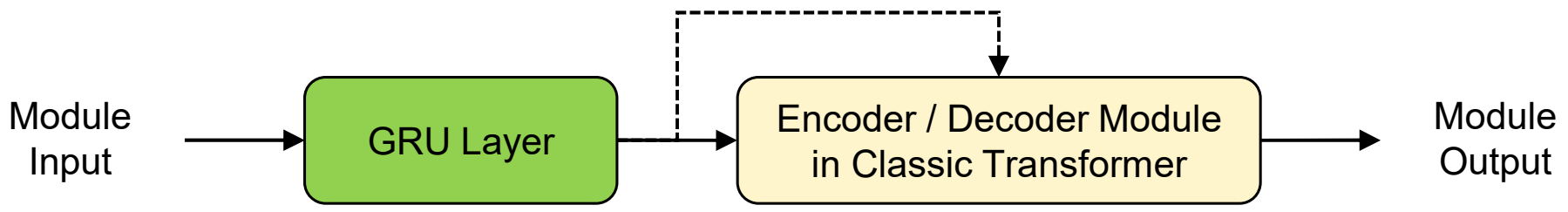

(c) Encoder and Decoder Unit in GA network

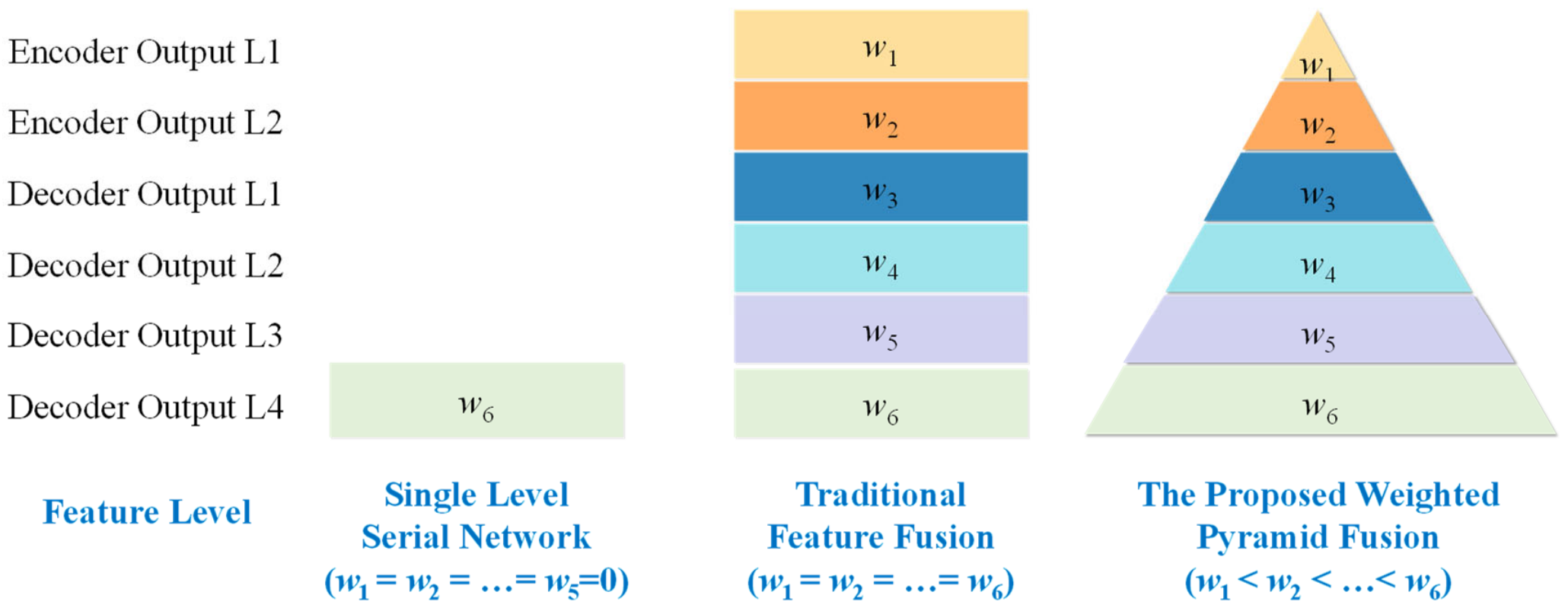

(d) Fusion mechanism of features

**Fig. 2** Comparison of modules in Transformer and GA network

Notably, the modules in Fig. 2(a) can be further partitioned, and the feature output of the submodules can be imported into the output module by adding shortcuts. In the following section of this study, the key settings (i.e., the number of layers and the method of adding shortcuts) are compared and analyzed.

## 2.3 Weighted stacked feature fusion mechanism

Traditional serial networks only consider the last level of feature when predicting the output (as shown in Figs. 1(a) and 2(d)). In addition, in previous studies, feature fusion is generally performed by (1) element-wise summarization, (2) concatenating specific dimensions, and (3) decreasing/increasing the spatial size together with element-wise summarization (when the spatial dimensions of the features are inconsistent). As illustrated in Fig. 2(d), these classic mechanisms consider different levels of features equally.

Although the fusion mechanisms mentioned above are straightforward, they disregard the differences in features at different levels. In many reasonably designed neural networks, the deeper the layer to which the features belong, the more significant the inherent relationship between the outputs and features, and vice versa. Although the features belonging to the shallow layers are essential, they are primarily used as the input of the subsequent layers. In other words, their positions are different.

Based on the analysis above, the different treatments of the features from different levels conform to the principle of neural network design. Therefore, a weighted feature fusion mechanism is proposed. This mechanism determines the weights of the features based on their levels, as shown in Fig. 2(d) and Equation (1). In terms of weight selection, referring to the typical principles of spatial size decay and channel number improvement in convolutional neural networks (e.g., Krizhevsky et al., 2012; He et al., 2016; Redmon and Farhadi, 2018), an exponential weight decay pattern (Equation (2)), is selected.

$$f_{Inte} = \frac{\sum_{i=1}^{n} f_i w_i}{\sum_{i=1}^{n} w_i} \tag{1}$$

$$w_i = \frac{1}{p^k} \; (p \geq 1.0) \tag{2}$$

where $f_{Inte}$ represents the integrated feature; $f_i$ represents the features of level $i$, such as the Encoder Outputs L1- L2 and Decoder Outputs L1-L4 shown in Fig. 2(a); $n$ represents the total number of features; $w_i$ represents the weight corresponding to $f_i$; $p$ is the weight decay factor, which will be discussed below; $k$ is the number of modules/layers between the layer (module) corresponding to $f_i$ and the output layer.

## 3. Dataset Establishment

All datasets used in this study are summarized in Table 1 (they can be downloaded through https://github.com/XYJ0904/Weighted-Pyramid-Stacked-Network).

### 3.1 Refined brace model

The refined finite element (FE) model of a brace component (Uriz and Mahin, 2008;

Huang, 2009) is selected. The model is established based on LS-DYNA using refined-mesh shell elements and a cyclic damage plasticity model. This model can simulate brace behaviors under complex loads, including material nonlinearity (yield, strengthening, and softening), geometric nonlinearity (large deformation, buckling), damage, and fracture propagation. Consequently, this model has been widely adopted (e.g., Li et al., 2013; Amiri et al., 2013; Moharrami and Koutromanos, 2017; Lu and Guan, 2021), and it is suitable as a representative benchmark.

Table 1 Summarization of the basic information of all datasets

| Case Name | Scale | Input/Output | Series Length | Dataset size | | | References |
|---|---|---|---|---|---|---|---|
| | | | | Train | Valid | Test | |
| OP-1<br>OP-2<br>OP-3<br>OP-4 | Material | Strain/<br>Stress | 1,000 | 4,000 | 1,000 | 1,000 | OpenSees wiki, 2021 |
| Huang<br>Huang-N | Component | Displacement/<br>Reaction force | 2,000 | 320 | 40 | 40 | Huang, 2009 |
| BoucWen | Structural | Ground Motion Acceleration/<br>Inter-story drift | 500 | 37 | 13 | 50 | Sun et al., 2013; Zhang et al., 2019 |
| MRFDBF | | | 1,000 | 47 | 20 | 481 | Dong et al., 2018; Zhang et al., 2019 |

Two hundred ground motion records with different characteristics are selected from the K-NET database (NIED, 2021). Their displacement sequences are obtained using the average acceleration integration method. In addition, based on several randomly selected sine waves, wave synthesis is conducted to construct 200 artificial displacement time series (Xu et al., 2022a). In the frequency domain, the adopted sine wave periods range from 0.2 s to 10 s, covering the most significant period in structural response analysis. The sine waves are amplitude by a factor $Bt$ ($B$ is a positive constant, and $t$ is time), which aims to increase the amplitude over time to ensure that the maximum amplitude appears in the late loading period. After the generation, the amplitudes of the displacement sequences are modified to the range between 76.2 and 88.9 mm (i.e., 3.0 and 3.5 inches, approximately 25-30 times the yield displacement of the brace) (Fig. 3(a)).

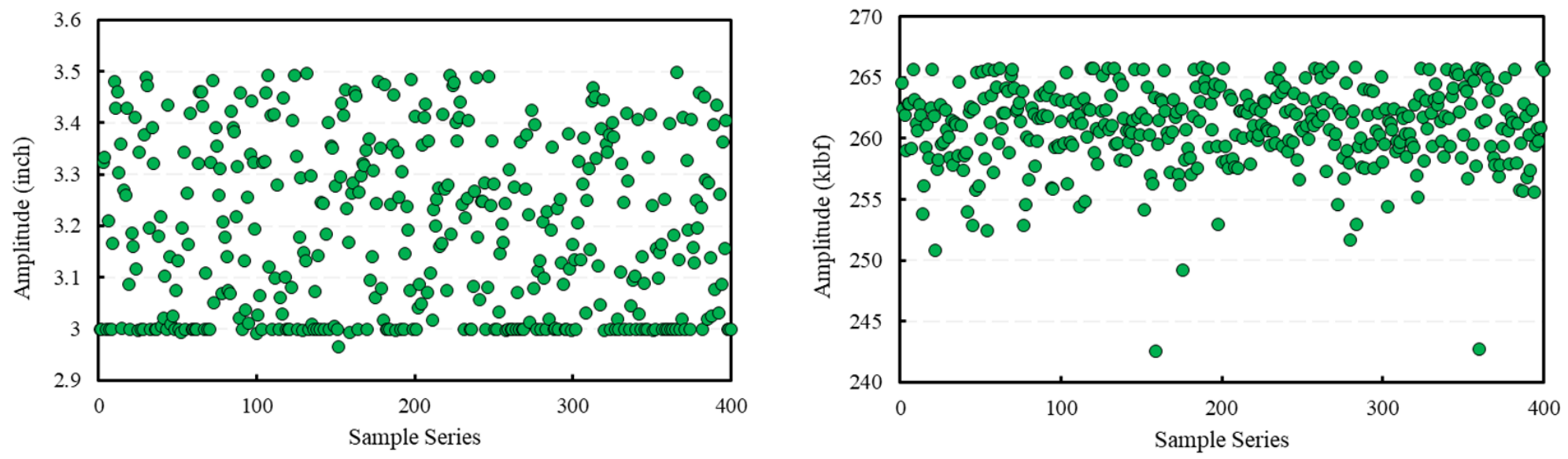

(a) Input amplitude (displacement) (b) Output amplitude (reaction force)

**Fig. 3** Typical amplitude distributions of inputs and outputs

Subsequently, the displacement sequences above are input into the LS-DYNA model, and 400 pairs of displacement-reaction force sequences are obtained. The simulation continues from the elastic stage to the failure stage, covering the entire hysteretic behavior. The amplitudes of the reaction force sequences are between 1081 and 1183 kN (243 and 267 klbf, Fig. 3(b)).

Based on the study of Zhang et al. (2019), the established dataset is normalized to the interval [-1, 1] through the min-max normalization. The original data without normalization are also retained to further verify the proposed method's robustness and analyze the effect of normalization. All samples are randomly segregated into three datasets containing 320, 40, and 40 samples, which are used as training, validation, and testing datasets, respectively. The two datasets with normalized and original data are denoted as cases "Huang-N" and "Huang," respectively.

3.2 Damped three-story frame

Dong et al. (2018) established a three-story frame model. This frame can be categorized into a lateral resisting system, a damping system, and a gravity load system. The lateral resisting system includes eight identical single-bay moment-resisting frames (MRFs), and the damping system comprises eight single-bay frames with nonlinear viscous dampers and the associated bracing (damped braced frame, DBF). The damping effect is significant in this case. Hereinafter, this case is referred to as "MRFDBF."

Based on the proposed model, Zhang et al. (2019) select 100 ground motions from the PEER NGA database (PEER, 2021). The ground motion amplitude adjustment and incremental dynamic analysis are then performed, which are widely adopted technologies in seismic response simulation. Subsequently, a dataset with 548 samples is established and normalized. In this dataset, each story's ground motion acceleration and inter-story drift are selected as the input and output, respectively. To prove the applicability of the redesigned network on small datasets, the dataset division method proposed by Zhang et al. (2019) is adopted. Only 47 samples are used for training, 20 samples are used for validation, and the remaining 481 samples are used for testing.

The amplitude distributions of the data are shown in Figs. 4(a-b). The output uniformly distributes in the range of 0.026 and 0.158 m, corresponding to an inter-story drift ratio (IDR) of 1/124 to 1/20. Such a response distribution encompasses the elastic and nonlinear phases of the frame.

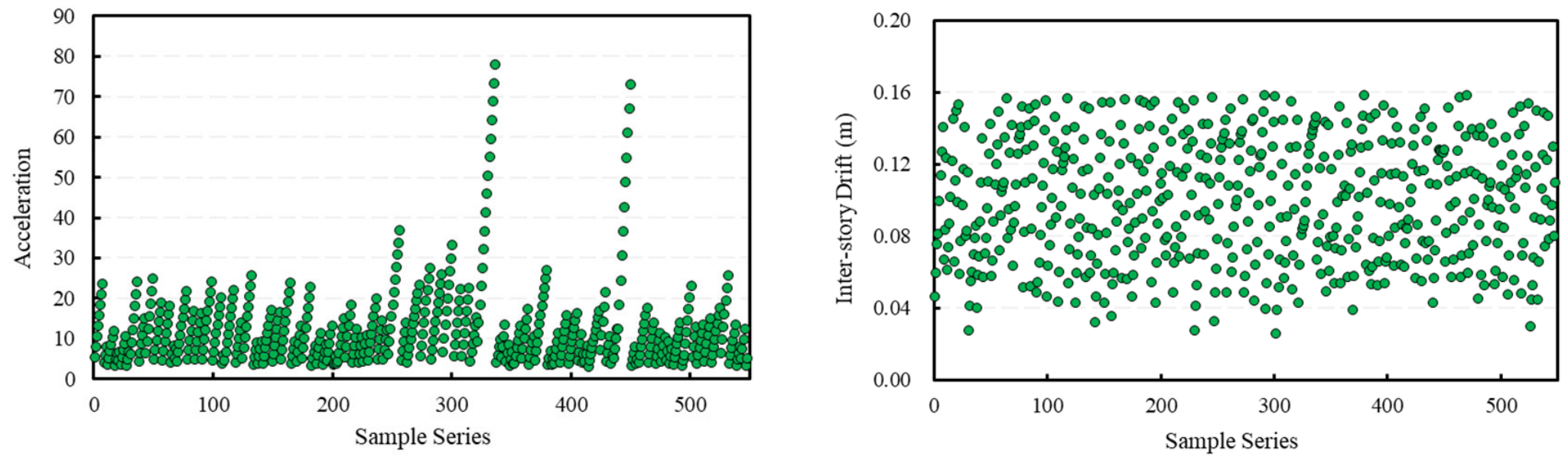

(a) Amplitude (ground motion) (b) Amplitude (inter-story drift)

**Fig. 4** Typical sample and amplitude distributions in case MRFDBF

### 3.3 Giuffré-Menegotto-Pinto material cases with different characteristics

Based on the Giuffré-Menegotto-Pinto material model in OpenSees, four combinations of recommended parameters (OpenSees wiki, 2021) are selected to construct datasets of material hysteretic behaviors with different characteristics. Specifically, ground motions are obtained based on the PEER NGA database (PEER, 2021) and the K-NET database (NIED, 2021) and pre-processed. Subsequently, the displacement sequences of those ground motions are used as the inputs to calculate the hysteretic behaviors of different materials. The strain and stress of the materials are selected as the input and output of the network, respectively. Finally, the datasets are normalized using the abovementioned method and randomly segregated into three groups containing 4,000, 1,000, and 1,000 samples, regarded as the training, validation, and testing datasets, respectively.

The distribution of the peak strain is shown in Fig. 5. The responses encompass both the elastic and elastoplastic stages. Therefore, the obtained datasets are representative. These four cases are denoted as OP-1 to OP-4, respectively, which include different hardening, softening, and deterioration levels.

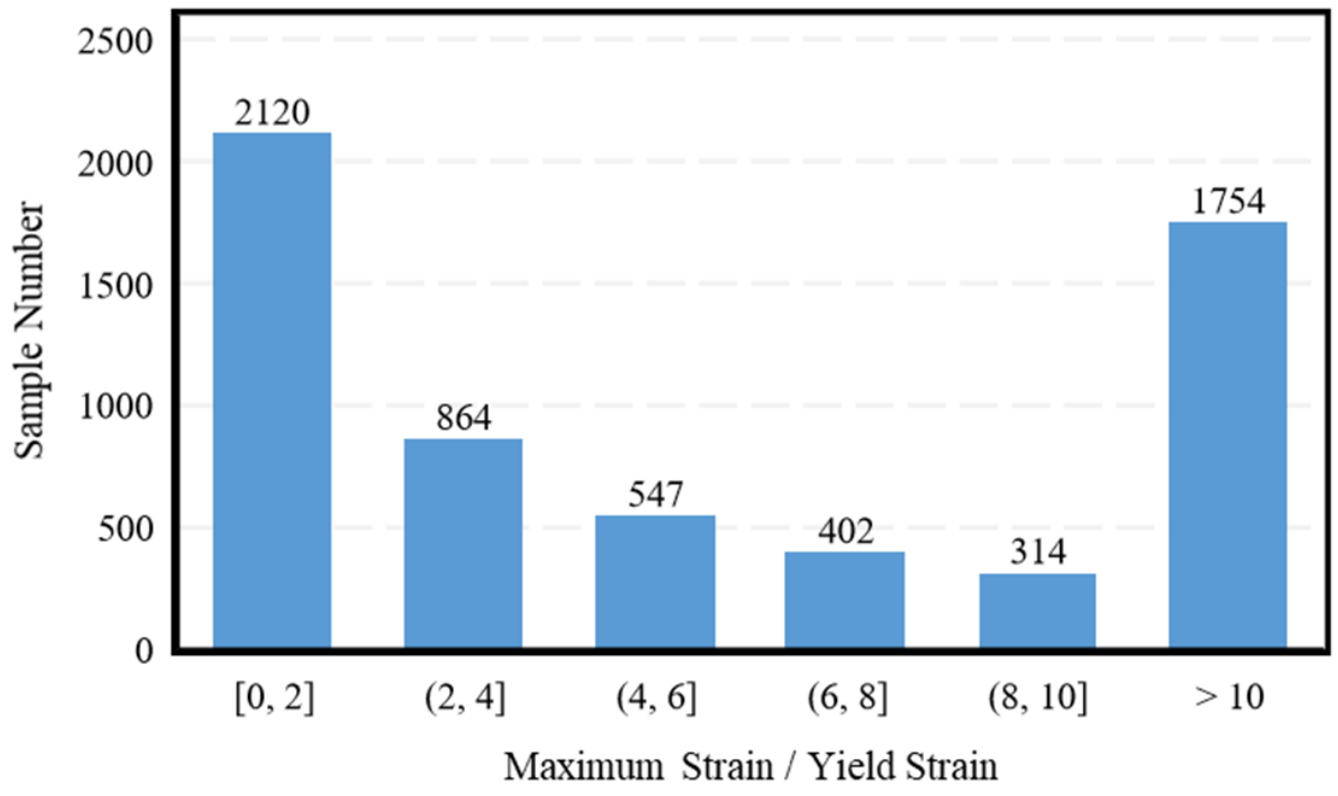

**Fig. 5** Typical strain distribution in cases OP-1 to OP-4

3.4 BoucWen case

The BoucWen model (Bouc, 1967; Wen, 1980) is a widely used nonlinear hysteretic model. The basic principle of this model is shown in Fig. 6(a). Zhang et al. (2019) set different key parameters and selected random band-limited white noise ground motion sequences with different magnitudes as inputs to construct a dataset containing 100 samples. A typical input sequence is shown in Fig. 6(b). The output is the inter-story drift of a five-degree-of-freedom system (Sun et al., 2013) based on the BoucWen model, and the normalization process is performed. Subsequently, the dataset is segregated into three groups containing 37, 13, and 50 samples, which are used as the training, validation, and testing datasets, respectively.

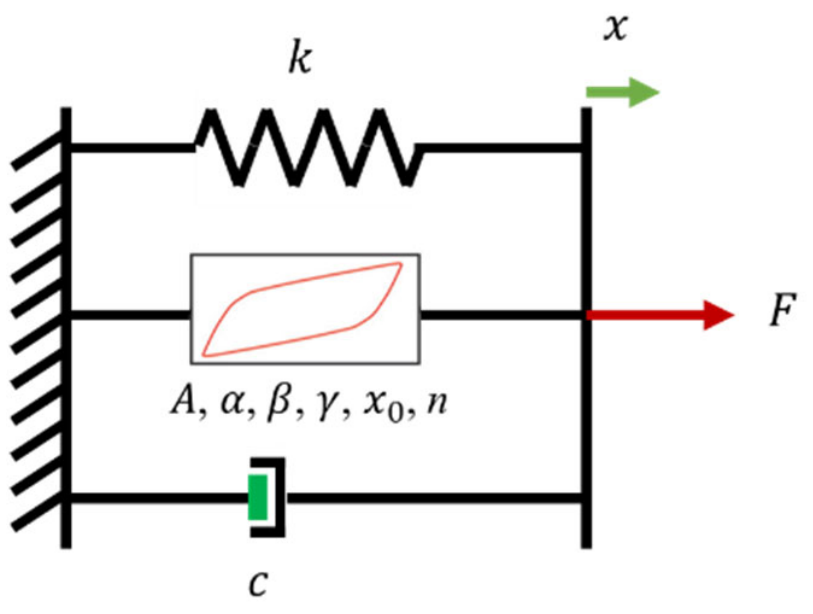

(a) BoucWen model

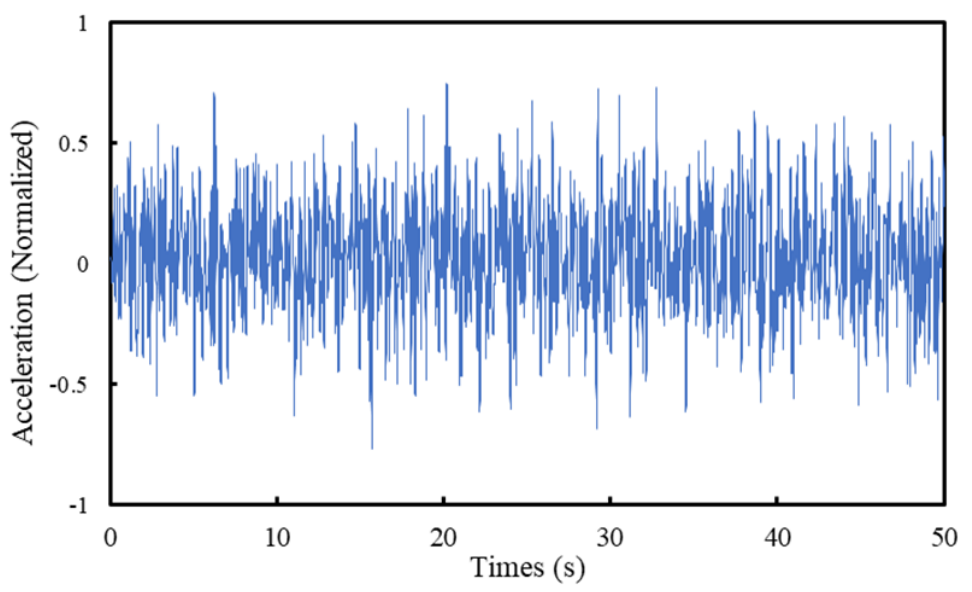

(b) Typical input sequence

**Fig. 6** The BoucWen model and typical input sequence

## 4. Discussion of Key Settings

4.1 Principle verification

As discussed in Section 2, the weighted pyramid network is designed based on the following two principles: (1) The positionings of features are varied, so different weights should be given in the feature fusion process; (2) Directly integrating multi-level features in the output module is beneficial for improving the final prediction performance.

To verify the rationality of these principles, a “separated” network is specially designed (as shown in Fig. 7). The design purpose of this network is to compare the predicted output based on different levels of features independently. This network thus consists of one feature extraction module and two independent prediction modules. The feature extraction module will extract multi-level features, which performs functions similar to the encoder and decoder modules in the conventional Transformer architecture.

The prediction modules are composed of several independent multi-layer perceptions (fully connected layers) that perform the final prediction based on the corresponding level of features.

It is necessary to cut off the gradient backpropagation between the feature extraction module and the prediction modules. Otherwise, the depth of the features (the number of layers between the selected hidden layer and final output) calculated through different paths will be conflicted, making it difficult to effectively define and distinguish the "level" of each feature. Therefore, in Fig. 7, light blue arrows represent connections between layers that allow both forward and backpropagation, while black arrows represent connections that only allow forward operation and do not allow gradient backpropagation.

In this section, cases Huang-N and MRFDBF are selected for verification. These two cases have been recognized in several follow-up studies. In addition, these relatively complex cases demonstrate consistency with practical applications. Therefore, the discussions based on these two cases are reliable. The model performances are shown in Fig. 8 and Table 2. The loss used in this section is the mean-square error (MSE).

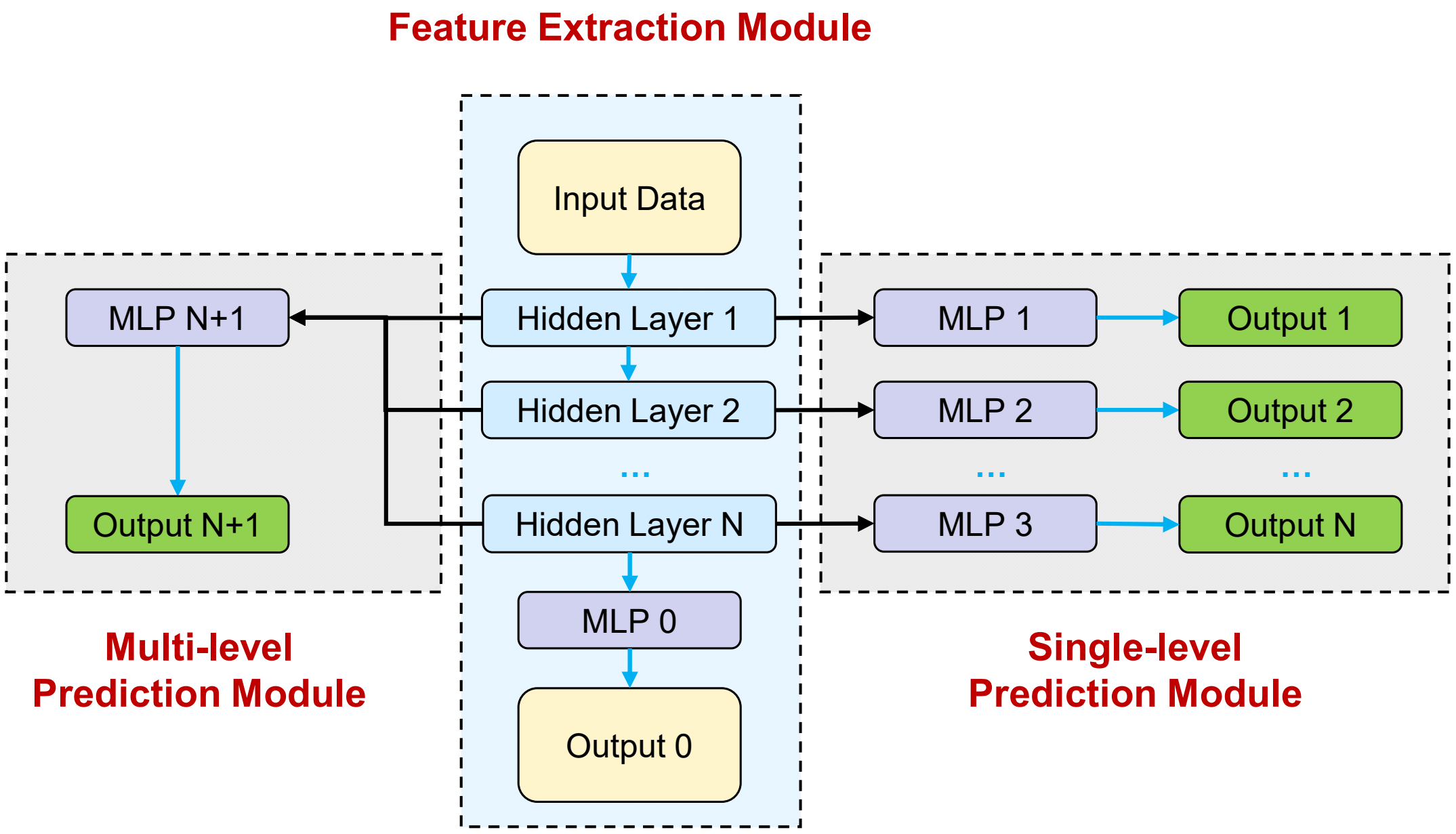

**Fig. 7** The "separated" network for principle verification

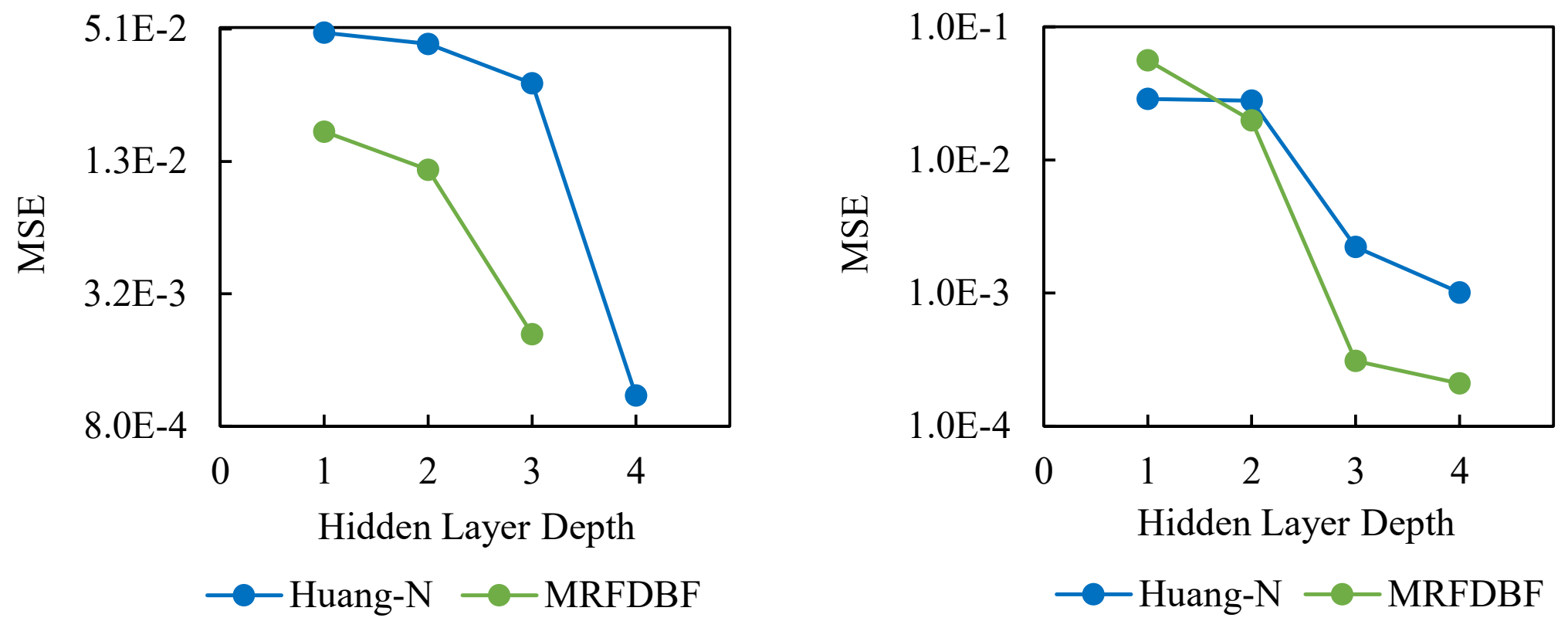

(a) Extraction module: LSTM (b) Extraction module: Transformer

**Fig. 8** Prediction performances based on different levels of features

Table 2 Comparison of the prediction performances

| Network for feature extraction | Case | Minimum MSE based on single-level features | Minimum MSE based on multi-level features |
|---|---|---|---|
| LSTM | Huang-N | $1.26\times10^{-3}$ | **$1.24\times10^{-3}$** |
| LSTM | MRFDBF | $4.44\times10^{-3}$ | **$4.22\times10^{-3}$** |
| Transformer | Huang-N | $2.61\times10^{-4}$ | **$2.21\times10^{-4}$** |
| Transformer | MRFDBF | $5.87\times10^{-4}$ | **$3.33\times10^{-4}$** |

From Fig. 8 and Table 2, it can be seen that:

(1) As the depth of the hidden layer increases, the prediction performance gradually improves, which confirms principle 1 mentioned above (features should be treated differently).

(2) The prediction results based on multi-level features are better than that based on single-level features, which confirms principle 2 (multi-level feature fusion in the output module is beneficial for prediction).

4.2 Comparison of key network settings

In the proposed weighted stacked pyramid architecture, various plans are available to add shortcuts. Meanwhile, different values of the weight decay factor $p$ (in Equation (2)) can be selected. Therefore, analysis and comparison will be carried out in this section. Some of the standard settings are as follows:

(1) The comparison and selection are conducted based on LSTM and Transformer.

As mentioned above, cases Huang, Huang-N, and MRFDBF are derived from existing literature and have been recognized in several follow-up studies. So they will be used for comparison.

(2) Owing to the variance of performances on different datasets, the "normalized loss" is proposed to quantify the relative advantages of different schemes, as shown in Equation (3).

$$Loss_N = \frac{Loss}{Loss_{max}} \tag{3}$$

where $Loss_N$ represents the normalized loss. $Loss$ and $Loss_{max}$ represent the absolute and maximum values of validation loss obtained for comparison, respectively.

4.2.1 Comparison of different LSTM architectures

The following four LSTM architectures are established to discuss the effects of the number of LSTM layers and how the shortcuts are added:

(1) LSTM-1: one LSTM layer and no shortcut is available

(2) LSTM-2: two LSTM layers and a shortcut is added after the first layer

(3) LSTM-3: three LSTM layers and shortcuts are added after the first two layers

(4) LSTM-4: three LSTM layers and a shortcut is added after the second layer

As shown in Fig. 9(a), LSTM-2 performs best; hence, it is selected for further comparison of different weight decay factors $p$. As shown in Fig. 9(b), when $p$ = 2.0, the network exhibits the best performance.

4.2.2 Comparison of different Transformer architectures

Five different Transformer architectures are designed, and the features imported into the output module (through multi-level shortcuts) in each architecture are shown below:

(1) Transformer-1 (TF-1): Decoder Outputs L3 and L4

(2) Transformer-2 (TF-2): Decoder Outputs L2 and L4

(3) Transformer-3 (TF-3): Decoder Outputs L2-L4

(4) Transformer-4 (TF-4): Decoder Outputs L1-L4

(5) Transformer-5 (TF-5): Decoder Outputs L1-L4; Encoder Outputs L1-L2

The comparison results are shown in Fig. 9(c), confirming that importing more levels of features into prediction is conducive to improving the network performance. The Transformer-5 (TF-5) network performs the best under the current circumstances; hence, it is selected for further comparison of the weight decay factor $p$, as shown in Fig. 9(d). When $p$ = 2.0, the network exhibits the best performance.

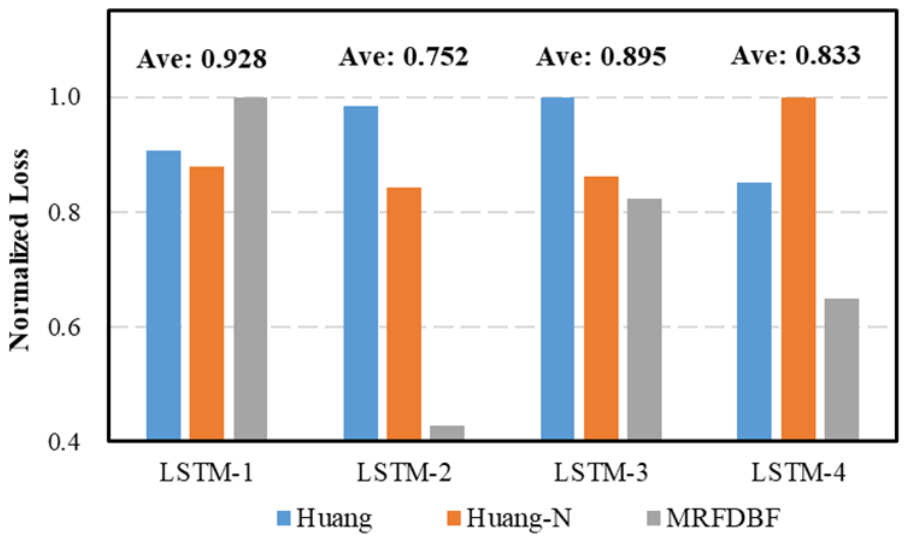

(a) Comparison of LSTM architectures

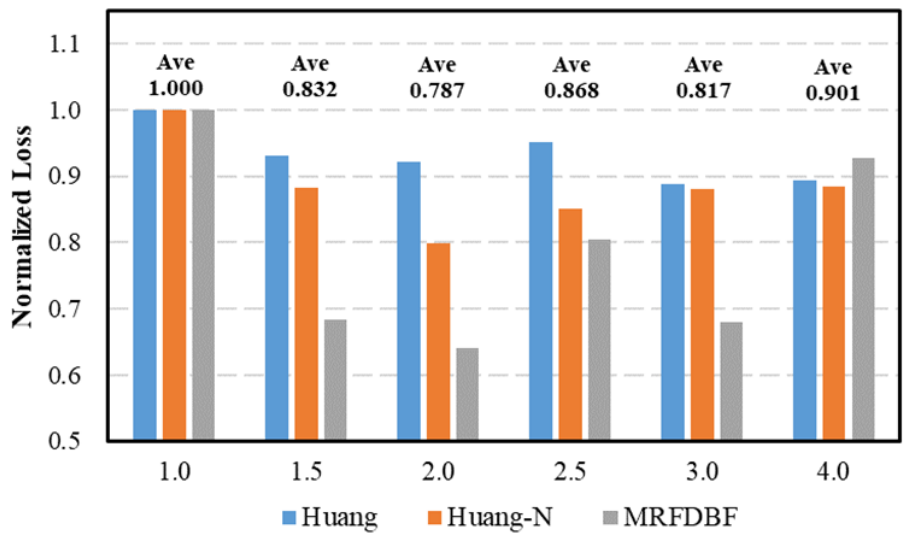

(b) Comparison of decay factors in LSTM

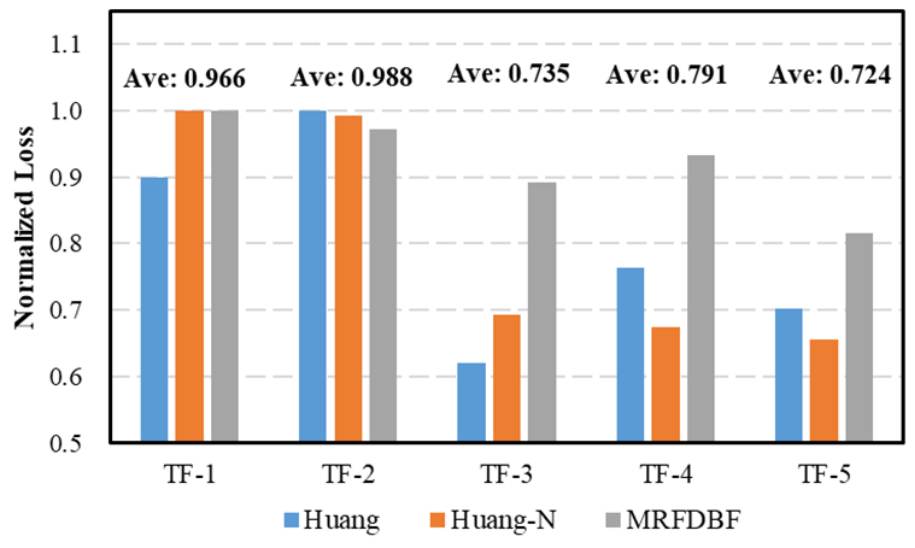

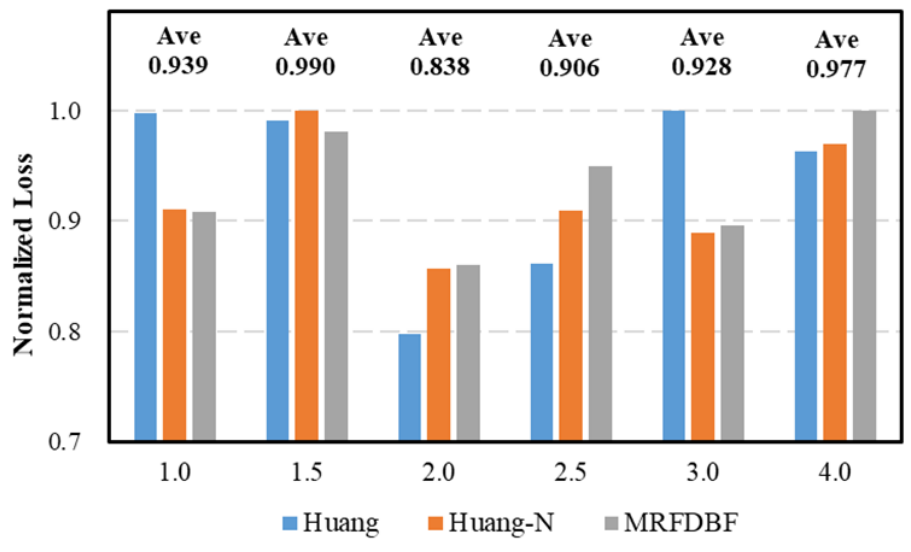

(c) Comparison of Transformer architectures

(d) Weight decay factors in Transformer

**Fig. 9** Comparison of different architectures and weight decay factors

## 5. Network Testing and Comparison

The redesigned networks (Pyramid-LSTM, Pyramid-Transformer, and Pyramid-GA) are tested on all eight datasets and compared with several commonly used networks in this section.

5.1 Selected networks for comparison

As mentioned in the introduction, commonly used neural networks in hysteretic behavior simulation include MLP, LSTM/GRU, attention-based networks (like Transformer), the combination of LSTM/GRU with attention mechanism, and CNN. Therefore, the following networks proposed and adopted in existing studies are selected for comparison, which cover the categories mentioned above.

(1) Classical LSTM, Transformer, MLP, and 1D-CNN

These networks are widely used models and thus will no longer be explained here.

(2) Physics-guided CNN (named “PhyCNN” hereafter)

The physics-informed neural network is an essential branch, so it is necessary to be considered in the comparison. PhyCNN, proposed by Zhang et al. (2020), is regarded as one of the representative studies of combining physics laws with neural networks and has been widely recognized in many follow-up studies. Additional constraints are added through the network architecture and loss function modification to improve the network performance.

(3) Recursive LSTM (named “Rec-LSTM” hereafter)

The Rec-LSTM network is an improved network proposed by Xu et al. (2022b) based on classical LSTM. This network divides the time series into multiple windows at certain intervals and predicts the response of the specified time step based on the information within a specific range (rather than using all historical time steps). This network is a representation of establishing local time-series correlation.

(4) Unrolled Attention Sequence-to-Sequence network (named “UA” hereafter)

The UA network is a response time-series prediction network proposed by Wang et al. (2020b). Many similar studies combine the LSTM/GRU with the attention mechanism (e.g., Li et al., 2021a, 2021b). Their network architectures are basically the same, so the UA network is selected to represent these studies.

## 5.2 Network comparison results

In this section, the training and testing of the networks and the comparisons of different architectures are carried out. The hyperparameters of all networks are thoroughly adjusted, and the best performances are adopted for comparison. The performance of each network is shown in Table 3. Fig. 10 shows the average normalized loss on the testing datasets of all eight cases.

Because certain physics laws (such as the differential relationships among the acceleration, velocity, and displacement sequences) are necessary for the training process of the PhyCNN, it has higher requirements on the fundamental dataset. The datasets constructed in Section 3 cannot fulfill the requirements. Therefore, the datasets and results provided by Zhang et al. (2020) (the authors of the PhyCNN paper) are adopted in comparison. Under the same physical constraints, the Pyramid-LSTM network is trained, and the results are shown in Table 4. It can be seen that the Pyramid-LSTM outperforms the PhyCNN on all three datasets. Due to the significant differences in data requirements, the PhyCNN network will not be further discussed.

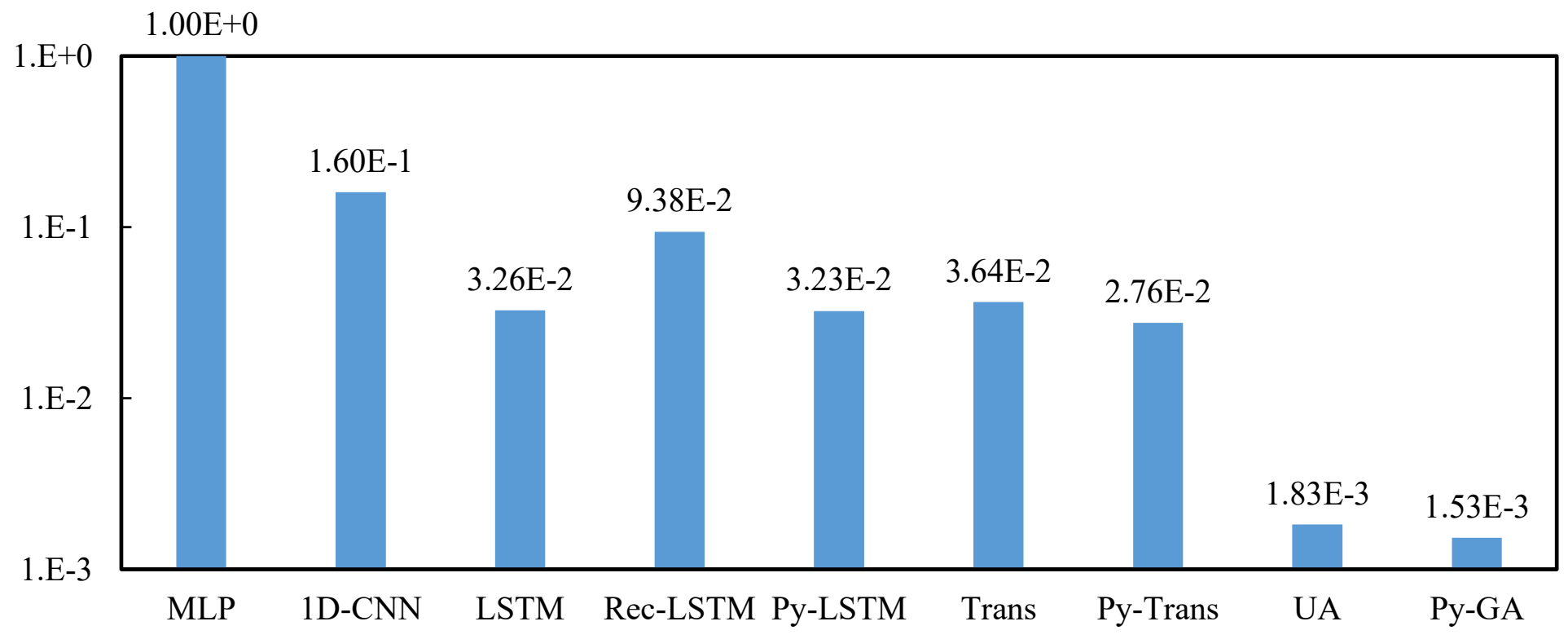

**Fig. 10** Average normalized loss of each network (testing datasets)

From Tables 3-4 and Fig. 10, the following conclusions can be drawn:

(1) The weighted stacked pyramid network architecture performs best in all cases. Compared with other existing networks, the testing MSEs are reduced by 34.7% on average.

(2) Using the normalized loss on the testing dataset as the indicator, the Pyramid-GA network performs the best, followed by the UA network, the Pyramid-Transformer, and the Pyramid-LSTM.

(3) The basic network architecture has a significant influence on the performance of the model. With the same basic architecture (for example, Pyramid-LSTM and LSTM have the same basic architecture), the redesigned weighted pyramid network outperforms the original network in 87.5 % of tasks. The proposed multi-level shortcuts and weighted feature fusion mechanism are versatile in different basic architectures.

To show the prediction deviations more intuitively, Fig. 11 provides the prediction results and corresponding ground truth of the three proposed networks in the testing dataset. For brevity, only the results of some cases are given here. In Fig. 11, the orange lines represent the ground truth, and the blue lines represent the predicted results. The prediction results based on the weighted pyramid networks agree well with the ground truth.

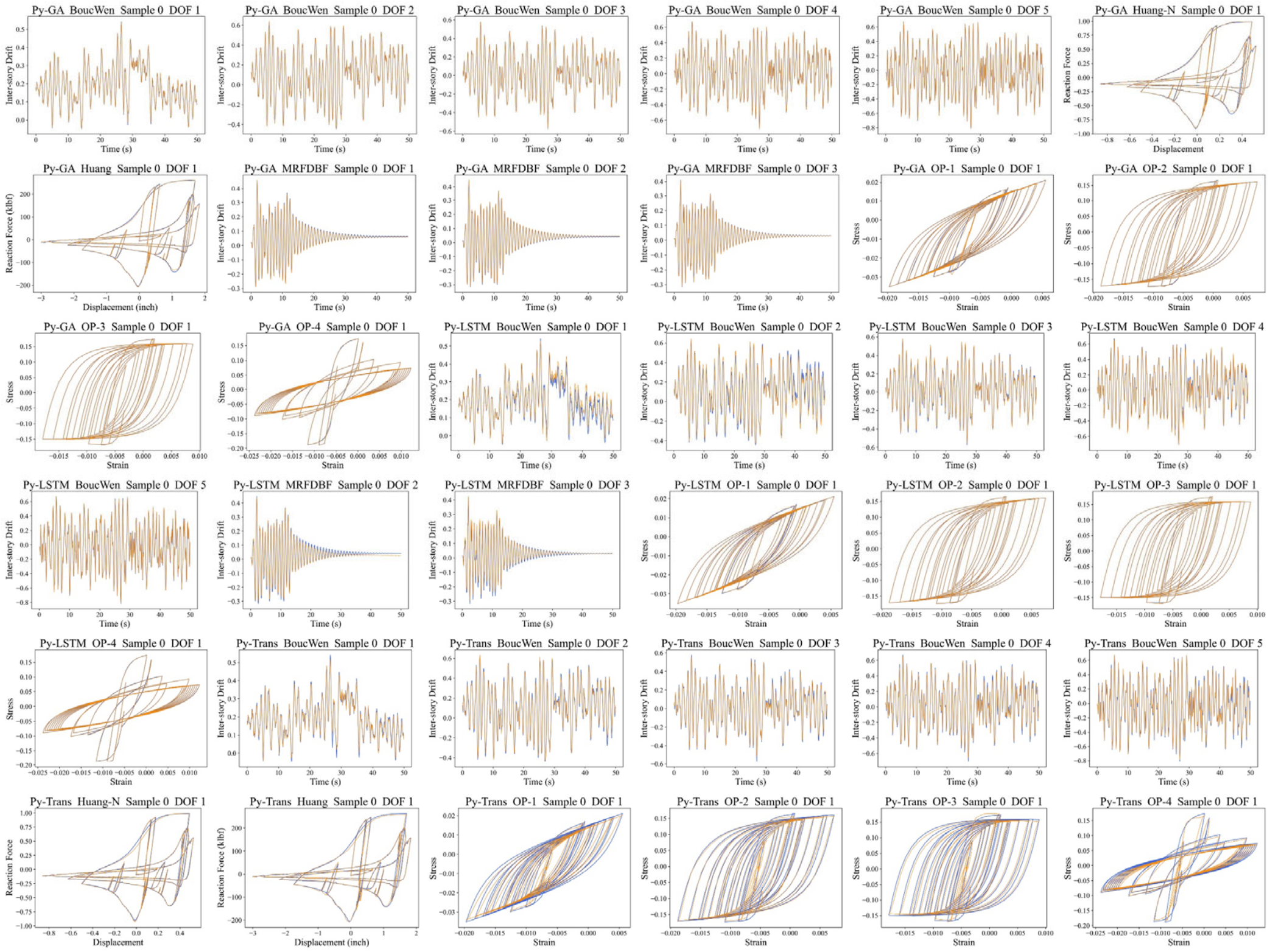

**Fig. 11** Typical prediction results of the weighted pyramid networks

Table 3 Network performances (Testing MSE)

| Case | LSTM | Transformer | MLP | UA | 1D-CNN | Rec-LSTM | Pyramid-LSTM | Pyramid-Transformer | Pyramid-GA |
|---|---|---|---|---|---|---|---|---|---|
| OP-1 | $2.73\times10^{-7}$ | $4.60\times10^{-6}$ | $5.81\times10^{-5}$ | $8.90\times10^{-8}$ | $6.88\times10^{-7}$ | $5.98\times10^{-6}$ | **$3.86\times10^{-8}$** | $3.23\times10^{-6}$ | $9.10\times10^{-8}$ |
| OP-2 | $1.08\times10^{-6}$ | $1.38\times10^{-4}$ | $5.57\times10^{-3}$ | $3.00\times10^{-6}$ | $2.48\times10^{-5}$ | $5.47\times10^{-5}$ | **$5.87\times10^{-7}$** | $1.17\times10^{-4}$ | $2.10\times10^{-6}$ |
| OP-3 | $1.28\times10^{-6}$ | $1.53\times10^{-4}$ | $6.26\times10^{-3}$ | $3.53\times10^{-6}$ | $3.34\times10^{-5}$ | $1.56\times10^{-4}$ | **$9.40\times10^{-7}$** | $1.44\times10^{-4}$ | $1.11\times10^{-6}$ |
| OP-4 | $3.08\times10^{-6}$ | $7.65\times10^{-5}$ | $1.70\times10^{-3}$ | $1.37\times10^{-6}$ | $1.09\times10^{-4}$ | $3.15\times10^{-5}$ | $3.86\times10^{-6}$ | $7.64\times10^{-5}$ | **$3.07\times10^{-7}$** |
| Huang-N | $7.63\times10^{-4}$ | $3.31\times10^{-4}$ | $7.79\times10^{-2}$ | $6.43\times10^{-5}$ | $8.35\times10^{-3}$ | $8.45\times10^{-4}$ | $7.47\times10^{-4}$ | $2.33\times10^{-4}$ | **$5.35\times10^{-5}$** |
| Huang | $5.85\times10^{1}$ | $1.68\times10^{1}$ | $4.78\times10^{3}$ | $3.47\times10^{0}$ | $4.65\times10^{2}$ | $3.63\times10^{1}$ | $4.17\times10^{1}$ | $1.04\times10^{1}$ | **$3.18\times10^{0}$** |
| MRFDBF | $4.98\times10^{-3}$ | $1.70\times10^{-3}$ | $2.58\times10^{-2}$ | $2.24\times10^{-4}$ | $1.71\times10^{-2}$ | $2.36\times10^{-3}$ | $4.90\times10^{-3}$ | $1.48\times10^{-3}$ | **$2.02\times10^{-4}$** |
| BoucWen | $2.99\times10^{-3}$ | $3.41\times10^{-3}$ | $7.81\times10^{-2}$ | $7.24\times10^{-5}$ | $2.52\times10^{-2}$ | $3.78\times10^{-2}$ | $3.64\times10^{-3}$ | $1.04\times10^{-3}$ | **$4.46\times10^{-5}$** |

Table 4 Comparison of the PhyCNN and Pyramid-LSTM

| Case | num_ag2utt | | num_ag2u | | exp_ag2utt | |
|---|---|---|---|---|---|---|
| | PhyCNN | Pyramid-LSTM | PhyCNN | Pyramid-LSTM | PhyCNN | Pyramid-LSTM |
| Loss 1 | $6.60\times10^{-4}$ | **$1.59\times10^{-4}$** | $9.43\times10^{-4}$ | **$2.19\times10^{-4}$** | $2.63\times10^{-6}$ | **$1.70\times10^{-6}$** |
| Loss 2 | $4.54\times10^{-1}$ | **$4.40\times10^{-1}$** | $2.41\times10^{-2}$ | **$7.11\times10^{-3}$** | $8.93\times10^{-1}$ | **$5.14\times10^{-1}$** |
| Loss 3 | - | - | $7.07\times10^{-1}$ | **$2.82\times10^{-1}$** | - | - |

Note: In case "num_ag2utt" and "exp_ag2utt", Loss 1 and Loss 2 represent the deviations of the predicted acceleration and displacement sequences, respectively. In case "num_ag2u", Loss 1, Loss 2, and Loss 3 represent the deviations of the predicted displacement, velocity, and acceleration sequences, respectively. Detailed information on these cases can be found in Zhang et al. (2020)

## 6 Experiment for Performance Variance Explanation

Compared to the Pyramid-LSTM network, in the Pyramid-GA network, the attention mechanism is introduced to improve model performance (Wang et al., 2020b). However, as shown in Table 3, the model prediction accuracy decreases unexpectedly in three cases (OP-1 to OP-3). The effect of the attention mechanism on model performance is inconsistent, and similar findings have been reported in other papers. (e.g., Li et al., 2021a, 2021b; Xu et al., 2022b). Existing studies did not provide explanations for these phenomena or corresponding verifications. Therefore, this section aims to create an innovative "memory experiment" to explain the performance variation partially.

Noting that the LSTM structure establishes implicit correlation among different time steps based on gated recurrent units (Hochreiter and Schmidhuber, 1997), this may result in superior short-term memory but poor long-term memory (Graves et al., 2014). The attention mechanism, on the other hand, is better at establishing global and long-term relationships through explicit attention calculation. Meanwhile, the long tail of attention values (Zhou et al., 2021) distracts network attention, potentially leading to relatively poor local and short-term memory ability.

To verify that the variances of long-term and short-term memory abilities significantly influence the model performance, the following "memory experiment" is designed. The output is the copy of the input, which is translated by $D$ time step later, as shown in Fig. 12. In this study, 15 datasets are established ($D$ = 10, 20, 30, 40, 50, 60, 70,

80, 90, 100, 200, 500, 1000, 1500, 1900). Each dataset comprises 1,000 samples with random start points, while the total sequence length (2,000) and interval $D$ are the same. 700, 200, and 100 samples are randomly divided into the training, validation, and testing datasets, respectively.

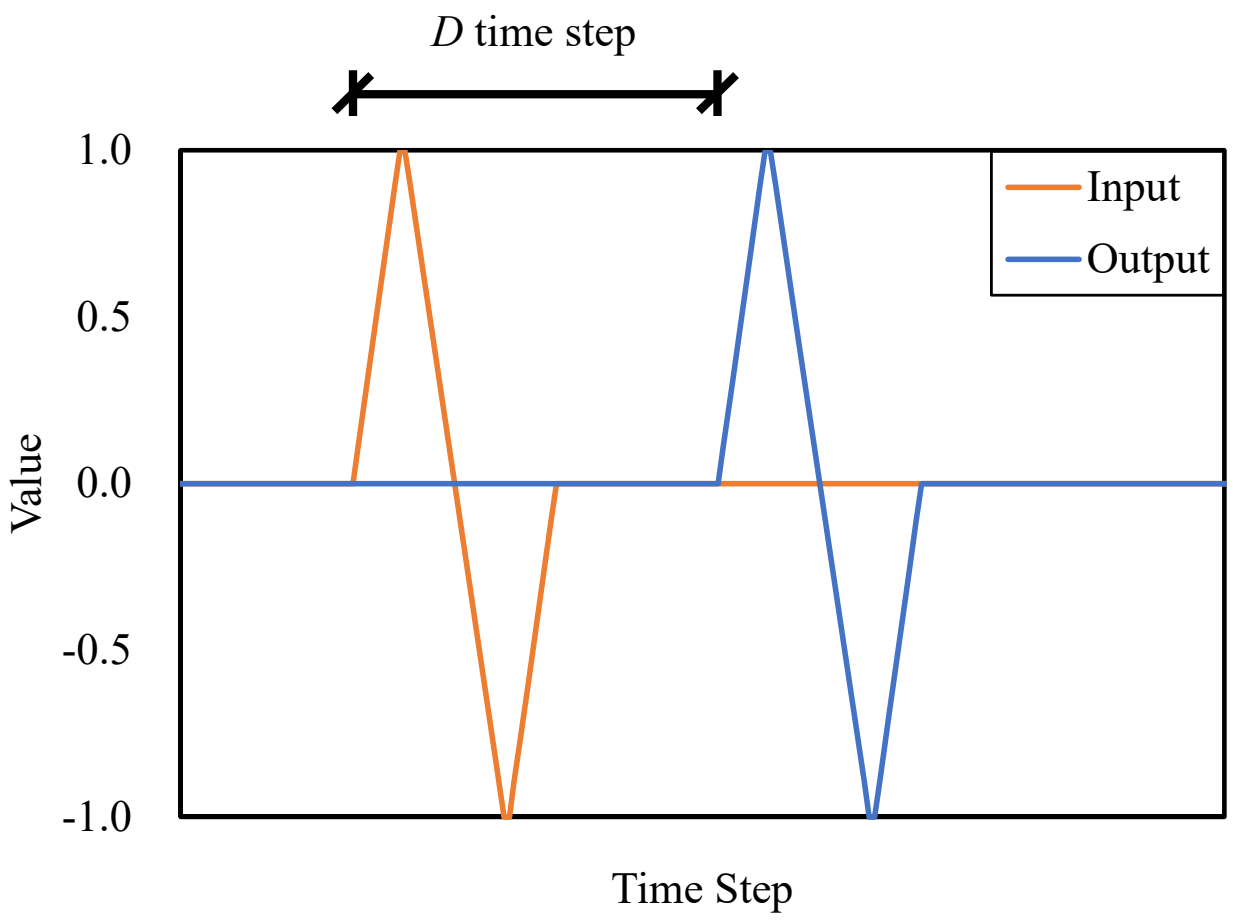

**Fig. 12** Typical sample in the memory experiment

The Pyramid-LSTM and Pyramid-Transformer (based on the attention mechanism) are trained and tested on these 15 datasets. Two groups of typical results with different hyperparameters (such as the number of layers and hidden dimensions) are shown in Fig. 13, and the trend in other groups is the same as that in Fig. 13.

The results show that the testing MSEs of the Pyramid-Transformer (attention-based) network are relatively consistent on datasets with different $D$, while a sudden change could be observed in the curves of Pyramid-LSTM between $D = 50$ and $D = 60$. For cases with $D \leq 50$, the MSEs of the Pyramid-LSTM network are orders of magnitude better than that of the Pyramid-Transformer. In contrast, diametrically opposite results are found in cases with $D \geq 60$.

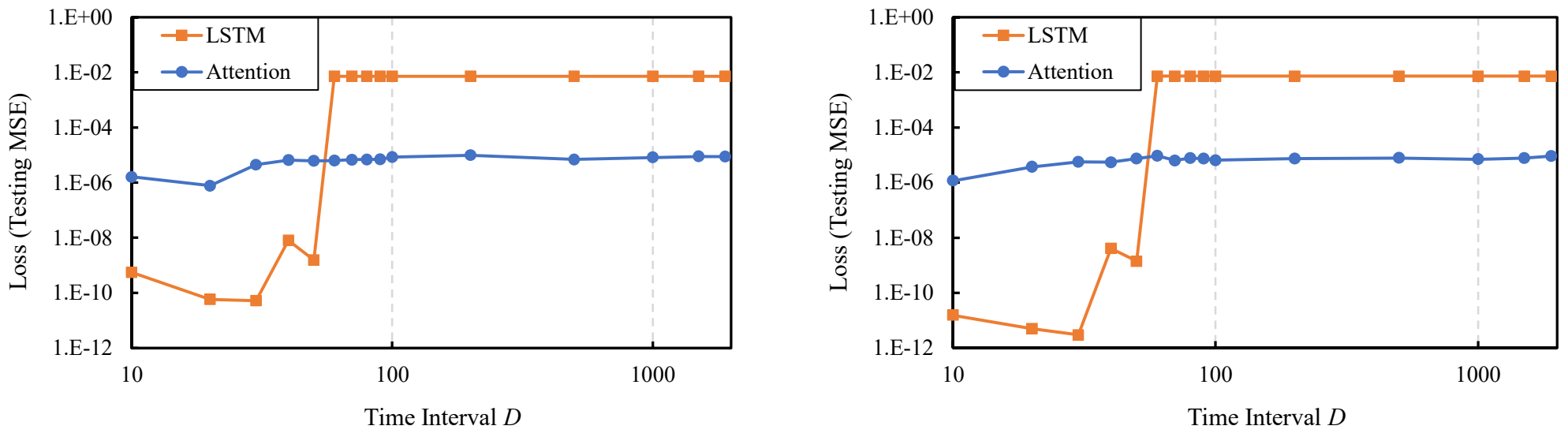

(a) Group 1 (b) Group 2

**Fig. 13** Model performances on datasets with different time intervals *D*

To eliminate the influence of random factors (such as parameter initialization) and hyperparameters, more Pyramid-LSTM networks are trained on critical datasets with $D = 50$ and $D = 60$. These networks have different hyperparameters, with the number of layers in [1, 4], learning rates in $[1 \times 10^{-4}, 5 \times 10^{-3}]$, and hidden dimensions in [100, 300]. The results are shown in Fig. 14(a), which confirms that the random factors and adjustment of hyperparameters do not affect the correctness of the conclusion. Meanwhile, typical results predicted by LSTM models are visualized in Fig. 14(b)-(c). When $D = 50$, the models translate the input successfully. Thus, the ground truth and prediction results are identical. However, when $D = 60$, the Pyramid-LSTM network does not capture any sequence features, so the predicted values at all time steps are zero (that is why the MSEs are 7.2 × $10^{-3}$ for all datasets with $D \geq 60$).

In summary, the Pyramid-LSTM network has better short-term memory ability yet poor long-term memory ability. The Pyramid-Transformer (attention-based) network has stable performances, while its short-term ability is worse than that of Pyramid-LSTM.

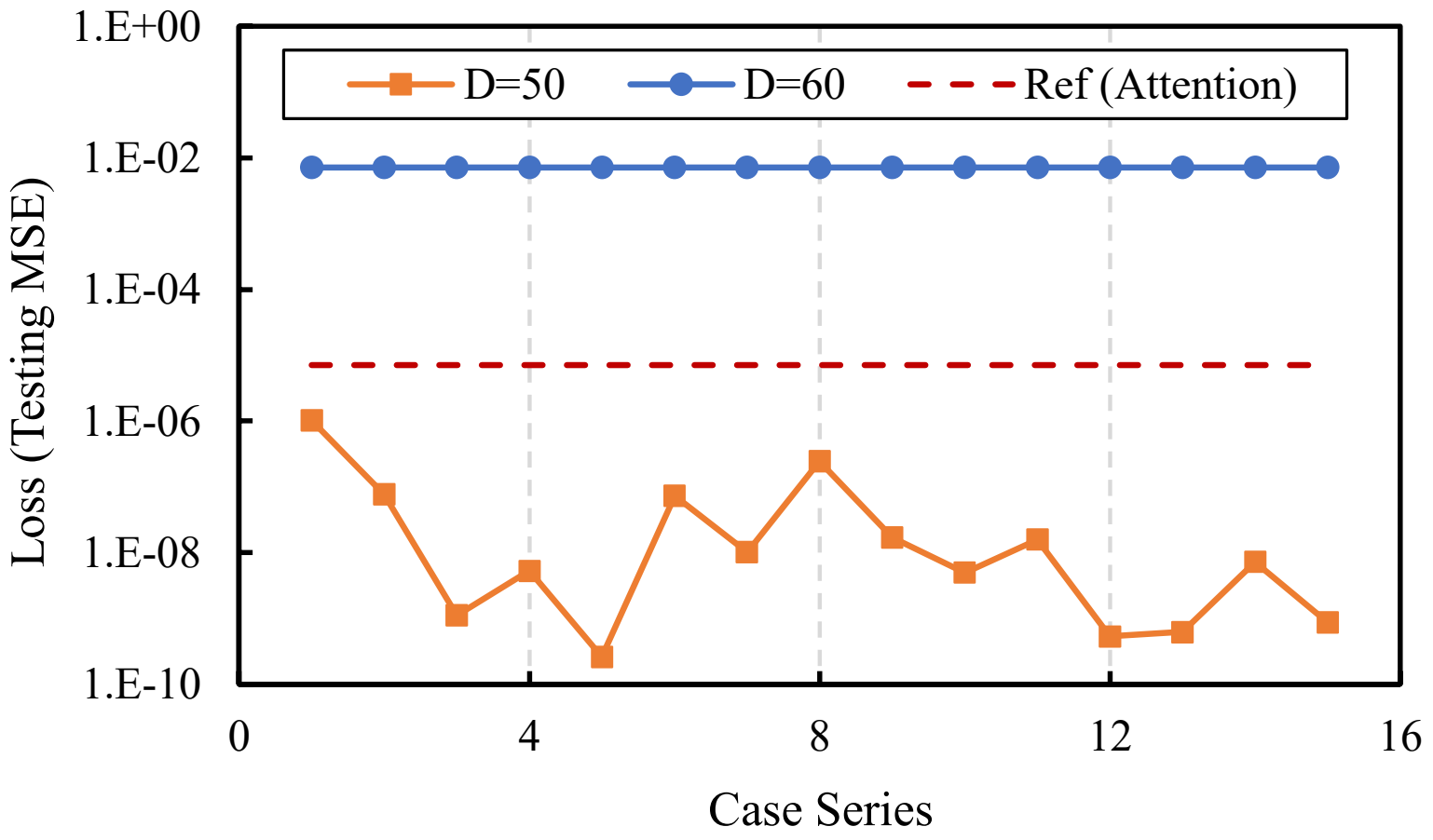

(a) Model performances on critical datasets with $D$ = 50 and $D$ = 60

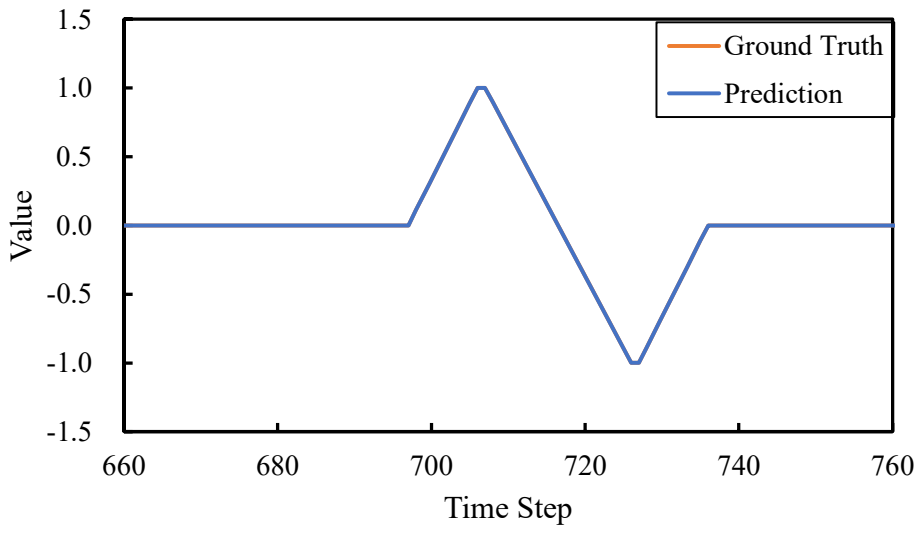

(b) Typical samples, $D$ = 50

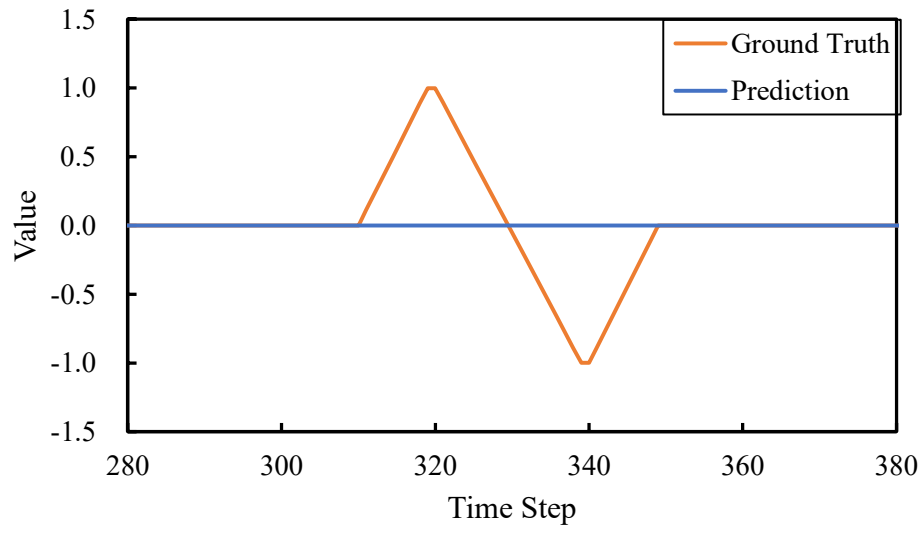

(c) Typical samples, $D$ = 60

**Fig. 14** Overall model performance and typical samples for critical time intervals

## 7. Conclusions

A weighted stacked pyramid network architecture is proposed based on existing network architectures to support hysteretic behavior simulations on various scales based on deep learning. Principle verification, datasets, network establishment, case study, and comparison among widely-adopted networks are carried out. The long and short-term memory abilities of different basic network architectures are compared and explained. The following conclusions are obtained:

(1) The proposed multi-level shortcut and weighted feature fusion strategy are reasonable. The testing MSEs of the proposed models are reduced by an average of 34.7% compared with existing widely-used networks. The prediction behaviors are consistent with the ground truth.

(2) Among all the network architectures discussed in this study, the proposed Pyramid-GA network performs the best, followed by the UA network, the Pyramid-Transformer, and the Pyramid-LSTM. For datasets dominated by short-term and long-term correlations, Pyramid-LSTM and Pyramid-GA (attention-based) networks are recommended, respectively.

(3) The weighted pyramid architecture proposed in this study is versatile. If the basic architecture is the same, the proposed networks can perform better in 87.5% of tasks.

**Data Availability Statement**

Datasets and codes used in this study can be downloaded through: https://github.com/XYJ0904/Weighted-Pyramid-Stacked-Network.

**Acknowledgment**

The authors are grateful for the financial support from the National Key R&D Program (No. 2019YFC1509305), the National Natural Science Foundation of China (No. 52238011), and the Tencent Foundation through the XPLORER PRIZE. This work is partly supported by the ZJU-UIUC Joint Research Center Project No. DREMES202001, and the Chaoyong Project from Haining Municipal. The authors also wish to thank Dr. Chen Wang and Ph.D. Candidate Shuang Song in Tsinghua University for their kind help.